\documentclass[10pt,journal,compsoc]{IEEEtran}
\ifCLASSOPTIONcompsoc
  \usepackage[nocompress]{cite}
\else
  \usepackage{cite}
\fi
\ifCLASSINFOpdf
\else
\fi
\usepackage[utf8]{inputenc} 
\usepackage[T1]{fontenc}    

\usepackage{epsfig}
\usepackage{graphicx}
\usepackage{amsmath}
\usepackage{amssymb}
\usepackage{enumitem}

\usepackage[pagebackref,breaklinks,colorlinks]{hyperref}
\usepackage[capitalize]{cleveref}
\crefname{section}{Sec.}{Secs.}
\crefname{table}{Table}{Tables}
\crefname{figure}{Fig.}{Figs.}

\usepackage{xcolor}
\usepackage{tcolorbox}

\usepackage[linesnumbered,algo2e,boxed]{algorithm2e}
\graphicspath{{figs/}}
\usepackage[figuresright]{rotating}
\usepackage{amsmath,amssymb,textcomp,subfigure,multirow,upgreek}
\usepackage{booktabs}
\usepackage{color,mdwlist}
\usepackage{tikz}
\usepackage[edges]{forest}
\definecolor{hidden-draw}{RGB}{20,68,106}
\definecolor{hidden-pink}{RGB}{255,245,247}

\makeatletter
\DeclareRobustCommand\onedot{\futurelet\@let@token\@onedot}
\def\@onedot{\ifx\@let@token.\else.\null\fi\xspace}

\def\eg{\emph{e.g}\onedot} 
\def\ie{\emph{i.e}\onedot}

\def\etal{\emph{et al}\onedot}
\makeatother

\usepackage{ragged2e}

\usepackage{review}

\usepackage{graphicx,fontawesome}
\graphicspath{{figs/}}

\usepackage{caption}
\usepackage{wrapfig}

\usepackage{enumitem}
\setitemize{noitemsep,topsep=0pt,parsep=0pt,partopsep=0pt}

\begin{document}
\title{A Survey on Multimodal Large Language Models}

\author{Shukang~Yin*,
    Chaoyou~Fu*$\dagger$,
    Sirui~Zhao*,
    Ke~Li, \\
    Xing~Sun,
    Tong~Xu,
    and~Enhong~Chen,~\IEEEmembership{Fellow,~IEEE}
	\IEEEcompsocitemizethanks{
 \IEEEcompsocthanksitem
$\dagger$Chaoyou Fu is the project leader.
\protect
 \IEEEcompsocthanksitem *Shukang Yin, Chaoyou Fu, and Sirui Zhao contribute equally.
 \protect
	    \IEEEcompsocthanksitem Shukang Yin, Sirui Zhao, Tong Xu, and Enhong Chen are with the Department of Data Science, University of Science and Technology of China, No.96, JinZhai Road Baohe District, Hefei, Anhui, 230026, China. E-mail: sirui@mail.ustc.edu.cn, cheneh@ustc.edu.cn
        \protect
        \IEEEcompsocthanksitem
        Chaoyou Fu, Ke Li, and Xing Sun are with the Tencent YouTu Lab, Shanghai 200233, China. E-mail: bradyfu24@gmail.com
        \protect
        }
	\thanks{Corresponding author: Chaoyou Fu, Sirui Zhao, and Enhong Chen.}
}

%
%

\markboth{IEEE Transactions on Pattern Analysis and Machine Intelligence}%
{Shell \MakeLowercase{\textit{et al.}}: Bare Demo of IEEEtran.cls for Computer Society Journals}
%



\IEEEtitleabstractindextext{%
\justify
\begin{abstract}
Recently, Multimodal Large Language Model (MLLM) represented by GPT-4V has been a new rising research hotspot, which uses powerful Large Language Models (LLMs) as a brain to perform multimodal tasks. 
The surprising emergent capabilities of MLLM, such as writing stories based on images and OCR-free math reasoning, are rare in traditional multimodal methods, suggesting a potential path to artificial general intelligence.
To this end, both academia and industry have endeavored to develop MLLMs that can compete with or even better than GPT-4V, pushing the limit of research at a surprising speed.
In this paper, we aim to trace and summarize the recent progress of MLLMs.
First of all, we present the basic formulation of MLLM and delineate its related concepts, including architecture, training strategy and data, as well as evaluation.
Then, we introduce research topics about how MLLMs can be extended to support more granularity, modalities, languages, and scenarios. 
We continue with multimodal hallucination and extended techniques, including Multimodal ICL (M-ICL), Multimodal CoT (M-CoT), and LLM-Aided Visual Reasoning (LAVR). 
To conclude the paper, we discuss existing challenges and point out promising research directions. 
In light of the fact that the era of MLLM has only just begun, we will keep updating this survey and hope it can inspire more research.
An associated GitHub link collecting the latest papers is available at \url{https://github.com/BradyFU/Awesome-Multimodal-Large-Language-Models}.
\end{abstract}

}

\maketitle

\IEEEdisplaynontitleabstractindextext

%
\IEEEpeerreviewmaketitle


%
%
%
%


\section{Introduction}
\label{sec:intro}

\IEEEPARstart{R}{ecent} years have seen the remarkable progress of LLMs~\cite{zhao2023survey,chatgpt, openai2023gpt4, vicuna, llama}. 
By scaling up data size and model size, these LLMs raise extraordinary emergent abilities, typically including instruction following ~\cite{llama, peng2023instruction}, In-Context Learning (ICL) ~\cite{brown2020language}, and Chain of Thought (CoT) ~\cite{wei2022chain}. 
Although LLMs have demonstrated surprising zero/few-shot reasoning performance on most Natural Language Processing (NLP) tasks, they are inherently ``blind'' to vision since they can only understand discrete text. 
Concurrently, Large Vision Models (LVMs) can see clearly~\cite{sam, shen2023aligning, dino, dinov2}, but commonly lag in reasoning. 

In light of this complementarity, LLM and LVM run towards each other, leading to the new field of Multimodal Large Language Model (MLLM).
Formally, it refers to the LLM-based model with the ability to receive, reason, and output with multimodal information.
Prior to MLLM, there have been a lot of works devoted to multimodality, which can be divided into discriminative~\cite{clip,li2021align,chen2020uniter} and generative~\cite{wang2022ofa,cho2021unifying,wang2021simvlm} paradigms.
CLIP~\cite{clip}, as a representative of the former, projects visual and textual information into a unified representation space, building a bridge for downstream multimodal tasks. 
In contrast, OFA~\cite{wang2022ofa} is a representative of the latter, which unifies multimodal tasks in a sequence-to-sequence manner. 
MLLM can be classified as the latter according to the sequence operation, but it manifests two representative traits compared with the traditional counterparts: 
(1) MLLM is based on LLM with billion-scale parameters, which is not available in previous models.
(2) MLLM uses new training paradigms to unleash its full potential, such as using multimodal instruction tuning~\cite{wei2021finetuned,llava} to encourage the model to follow new instructions.
Armed with the two traits, MLLM exhibits new capabilities, such as writing website code based on images~\cite{minigpt-4}, understanding the deep meaning of a meme~\cite{mm-react}, and OCR-free math reasoning~\cite{palm-e}.

\begin{figure*}[t]
    \centering
    \includegraphics[width=\linewidth]{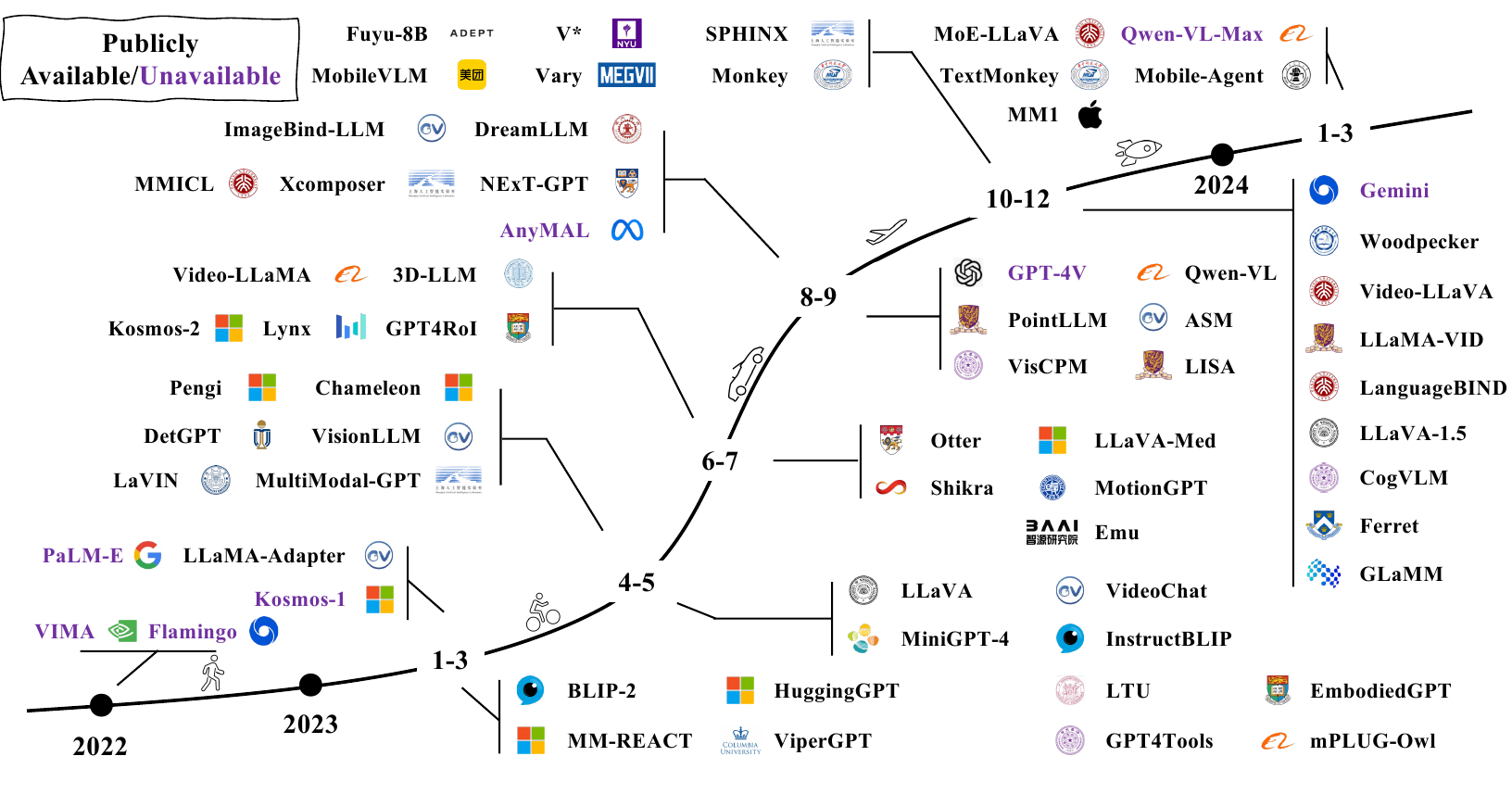}
    \caption{A timeline of representative MLLMs. We are witnessing rapid growth in this field. More works can be found in our released GitHub page, which is updated daily.}
    \label{fig:timeline}
\end{figure*}

Ever since the release of GPT-4~\cite{openai2023gpt4}, there has been a research frenzy over MLLMs because of the amazing multimodal examples it shows. 
Rapid development is fueled by efforts from both academia and industry. 
Preliminary research on MLLMs focuses on text content generation grounded in text prompts and image~\cite{llava,awadalla2023openflamingo}/video~\cite{videochat,video-llama}/audio~\cite{deshmukh2024pengi}.
Subsequent works have expanded the capabilities or the usage scenarios, including: 
(1) Better granularity support. Finer control on user prompts is developed to support specific regions through boxes~\cite{chen2023shikra} or a certain object through a click~\cite{yuan2023osprey}. 
(2) Enhanced support on input and output modalities~\cite{han2023imagebind, moon2023anymal}, such as image, video, audio, and point cloud. 
Besides input, projects like NExT-GPT~\cite{wu2023next} further support output in different modalities. 
(3) Improved language support. Efforts have been made to extend the success of MLLMs to other languages (\eg Chinese) with relatively limited training corpus~\cite{hu2023large, bai2023qwen}. 
(4) Extension to more realms and usage scenarios. Some studies transfer the strong capabilities of MLLMs to other domains such as medical image understanding~\cite{llava-med,moor2023med,pmc-vqa} and document parsing~\cite{ye2023mplug,liu2024textmonkey,hu2023mplug}. Moreover, multimodal agents are developed to assist in real-world interaction, \eg embodied agents~\cite{huang2023embodied,peng2023kosmos} and GUI agents~\cite{yang2023appagent,hong2023cogagent, wang2024mobile}. An MLLM timeline is illustrated in~\cref{fig:timeline}.

In view of such rapid progress and the promising results of this field, we write this survey to provide researchers with a grasp of the basic idea, main method, and current progress of MLLMs. Note that we mainly focus on visual and language modalities, but also include works involving other modalities like video and audio. 
Specifically, we cover the most important aspects of MLLMs with corresponding summaries and open a GitHub page that would be updated in real time. 
To the best of our knowledge, this is the first survey on MLLM.

The following parts of the survey are structured as such: the survey starts with a comprehensive review of the essential aspects of MLLMs, including 
(1) Mainstream architectures (\S \ref{sec:mllm_arch});
(2) A full recipe of training strategy and data (\S \ref{sec:mllm_train}); (3) Common practices of performance evaluation (\S \ref{sec:mllm_eval}). 
Then, we delve into a deeper discussion on some important topics about MLLMs, each focusing on a main problem: (1) What aspects can be further improved or extended (\S \ref{sec:extensions})? (2) How to relieve the multimodal hallucination issue (\S \ref{sec:hallu})?
The survey continues with the introduction of three key techniques (\S \ref{sec:tech}), each specialized in a specific scenario: M-ICL (\S \ref{sec:micl}) is an effective technique commonly used at the inference stage to boost few-shot performance. Another important technique is M-CoT (\S \ref{sec:mcot}), which is typically used in complex reasoning tasks. Afterward, we delineate a general idea to develop LLM-based systems to solve composite reasoning tasks or to address common user queries (\S \ref{sec:vr}).
Finally, we finish our survey with a summary and potential research directions.

\section{Architecture}
\label{sec:mllm_arch}
\begin{figure}[!htpb]
    \centering
    \includegraphics[width=\linewidth]{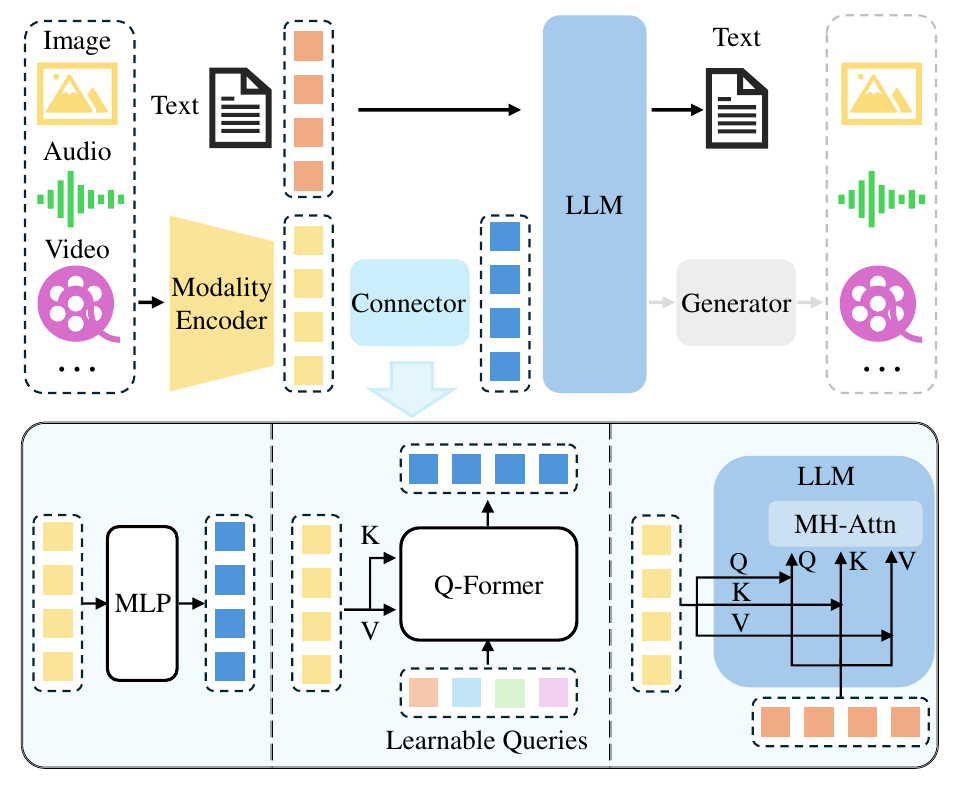}
    \caption{An illustration of typical MLLM architecture. It includes an encoder, a connector, and a LLM. An optional generator can be attached to the LLM to generate more modalities besides text. The encoder takes in images, audios or videos and outputs features, which are processed by the connector so that the LLM can better understand. There are broadly three types of connectors: projection-based, query-based, and fusion-based connectors. The former two types adopt token-level fusion, processing features into tokens to be sent along with text tokens, while the last type enables a feature-level fusion inside the LLM.}
    \label{fig:arch}
\end{figure}

A typical MLLM can be abstracted into three modules, \ie a pre-trained modality encoder, a pre-trained LLM, and a modality interface to connect them. Drawing an analogy to humans, modality encoders such as image/audio encoders are human eyes/ears that receive and pre-process optical/acoustic signals, while LLMs are like human brains that understand and reason with the processed signals. 
In between, the modality interface serves to align different modalities. 
Some MLLMs also include a generator to output other modalities apart from text.
A diagram of the architecture is plotted in~\cref{fig:arch}. In this section, we introduce each module in sequence. 

\begin{table*}[!t]
\centering
\caption{A summary of commonly used image encoders.}
\begin{tabular}{lcccc}
\toprule
\textbf{Variants} & \textbf{Pretraining Corpus}            & \textbf{Resolution} & \textbf{Samples (B)} & \textbf{Parameter Size (M)} \\ \midrule
OpenCLIP-ConvNext-L~\cite{cherti2023reproducible}     & LAION-2B     & 320 & 29 & 197.4  \\ 
CLIP-ViT-L/14~\cite{clip}           & OpenAI's WIT & 224/336 & 13 & 304.0  \\ 
EVA-CLIP-ViT-G/14~\cite{sun2023eva} & \multicolumn{1}{l}{LAION-2B,COYO-700M} & 224               & 11                         & 1000.0                      \\
 
OpenCLIP-ViT-G/14~\cite{cherti2023reproducible}       & LAION-2B     & 224 & 34 & 1012.7 \\ 
OpenCLIP-ViT-bigG/14~\cite{cherti2023reproducible}    & LAION-2B     & 224 & 34 & 1844.9 \\  \bottomrule
\end{tabular}
\label{tab:encoder}
\end{table*}

\subsection{Modality encoder}
The encoders compress raw information, such as images or audio, into a more compact representation.
Rather than training from scratch, a common approach is to use a pre-trained encoder that has been aligned to other modalities. For example, CLIP~\cite{clip} incorporates a visual encoder semantically aligned with the text through large-scale pre-training on image-text pairs. Therefore, it is easier to use such initially pre-aligned encoders to align with LLMs through alignment pre-training (see \S \ref{sec:train_pre_train}). 

The series of commonly used image encoders are summarized in~\cref{tab:encoder}. Apart from vanilla CLIP image encoders~\cite{clip}, some works also explore using other variants. For example, MiniGPT-4~\cite{minigpt-4} adopts an EVA-CLIP~\cite{fang2023eva,sun2023eva} (ViT-G/14) encoder, which is trained with improved training techniques.
In contrast, Osprey~\cite{yuan2023osprey} introduces a convolution-based ConvNext-L encoder~\cite{cherti2023reproducible} to utilize higher resolution and multi-level features.  
Some works also explore encoder-free architecture. For instance, the image patches of Fuyu-8b~\cite{fuyu-8b} are directly projected before sending to LLMs. Thus, the model naturally supports flexible image resolution input.

When choosing encoders, one often considers factors like resolution, parameter size, and pretraining corpus. 
Notably, many works have empirically verified that using higher resolution can achieve remarkable performance gains~\cite{bai2023qwen,liu2023improved,li2023monkey,mckinzie2024mm1}. The approaches for scaling up input resolution can be categorized into direct scaling and patch-division methods. 
The direct scaling way inputs images of higher resolutions to the encoder, which often involves further tuning the encoder~\cite{bai2023qwen} or replacing a pre-trained encoder with higher resolution~\cite{liu2023improved}. Similarly, CogAgent~\cite{hong2023cogagent} uses a dual-encoder mechanism, where two encoders process high and low-resolution images, respectively. High-resolution features are injected into the low-resolution branch through cross-attention.
Patch-division methods cut a high-resolution image into patches and reuse the low-resolution encoder. For example, Monkey~\cite{li2023monkey} and SPHINX~\cite{lin2023sphinx} divide a large image into smaller patches and send sub-images together with a downsampled high-resolution image to the image encoder, where the sub-images and the low-resolution image capture local and global features, respectively. 
In contrast, parameter size and training data composition are of less importance compared with input resolution, found by empirical studies~\cite{mckinzie2024mm1}.

Similar encoders are also available for other modalities. For example, Pengi~\cite{deshmukh2024pengi} uses CLAP~\cite{elizalde2023clap} model as the audio encoder. ImageBind-LLM~\cite{han2023imagebind} uses the ImageBind~\cite{girdhar2023imagebind} encoder, which supports encoding image, text, audio, depth, thermal, and Inertial Measurement Unit (IMU) data. Equipped with the strong encoder, ImageBind-LLM can respond to the input of multiple modalities.

\begin{table*}[!t]
\centering
\caption{A summary of commonly used open-sourced LLMs. en, zh, fr, and de stand for English, Chinese, French, and German, respectively.}
\label{tab:llm}
\begin{tabular}{lccccc}
\toprule
\textbf{Model} & \textbf{Release Date} & \textbf{Pretrain Data Scale} & \textbf{Parameter Size  (B)} & \textbf{Language Support} & \textbf{Architecture} \\
\midrule
Flan-T5-XL/XXL~\cite{chung2022scaling} & Oct-2022 & -           & 3/ 11           & en, fr, de & Encoder-Decoder \\
LLaMA~\cite{llama}          & Feb-2023 & 1.4T tokens & 7/ 13/ 33/ 65   & en         & Causal Decoder  \\
Vicuna~\cite{vicuna}         & Mar-2023 & 1.4T tokens & 7/ 13/ 33       & en         & Causal Decoder  \\
LLaMA-2~\cite{llama-2}        & Jul-2023 & 2T tokens   & 7/ 13/ 70       & en         & Causal Decoder  \\
Qwen~\cite{qwen-lm}           & Sep-2023 & 3T tokens   & 1.8 / 7/ 14/ 72 & en, zh     & Causal Decoder  \\
\bottomrule
\end{tabular}
\end{table*}

\subsection{Pre-trained LLM}
Instead of training an LLM from scratch, it is more efficient and practical to start with a pre-trained one. Through tremendous pre-training on web corpus, LLMs have been embedded with rich world knowledge, and demonstrate strong generalization and reasoning capabilities. 

We summarize the commonly used and publicly available LLMs in~\cref{tab:llm}. Notably, most LLMs fall in the causal decoder category, following GPT-3~\cite{brown2020language}. Among them, Flan-T5~\cite{chung2022scaling} series are relatively early LLMs used in works like BLIP-2~\cite{li2023blip} and InstructBLIP~\cite{instructblip}. LLaMA series~\cite{llama,llama-2} and Vicuna family~\cite{vicuna} are representative open-sourced LLMs that have attracted much academic attention. Since the two LLMs are predominantly pre-trained on English corpus, they are limited in multi-language support, such as Chinese. In contrast, Qwen~\cite{qwen-lm} is a bilingual LLM that supports Chinese and English well. 

It should be noted that scaling up the parameter size of LLMs also brings additional gains, similar to the case of increasing input resolution. Specifically, Liu \etal \cite{liu2023improved, liu2024llavanext} find that simply scaling up LLM from 7B to 13B brings comprehensive improvement on various benchmarks. Furthermore, when using a 34B LLM, the model shows emergent zero-shot Chinese capability, given that only English multimodal data are used during training. Lu~\etal~\cite{lu2023empirical} see a similar phenomenon by scaling up LLMs from 13B to 35B and 65B/70B, where the larger model size brings consistent gains on benchmarks specifically designed for MLLMs. 
There are also works that use smaller LLMs to facilitate deployment on mobile devices. For example, MobileVLM series~\cite{chu2023mobilevlm,chu2024mobilevlm} use downscaled LLaMA~\cite{llama} (termed as MobileLLaMA 1.4B/2.7B), enabling efficient inference on mobile processors.

Recently, explorations of Mixture of Experts (MoE) architecture for LLMs have garnered rising attention~\cite{shen2023mixture,jiang2024mixtral,fedus2022switch}. Compared with dense models, the sparse architecture enables scaling up total parameter size without increasing computational cost, by selective activation of the parameters. Empirically, MM1~\cite{mckinzie2024mm1} and MoE-LLaVA~\cite{lin2024moe} find that MoE implementation achieves better performance than the dense counterpart on almost all the benchmarks.

\subsection{Modality interface}
\label{sec:arch_modality_interface}
Since LLMs can only perceive text, bridging the gap between natural language and other modalities is necessary. 
However, it would be costly to train a large multimodal model in an end-to-end manner.  
A more practical way is to introduce a learnable connector between the pre-trained visual encoder and LLM. 
The other approach is to translate images into languages with the help of expert models, and then send the language to LLM.

\noindent \textbf{Learnable Connector.}
It is responsible for bridging the gap between different modalities.
Specifically, the module projects information into the space that LLM can understand efficiently. 
Based on how multimodal information is fused, there are broadly two ways to implement such interfaces, \ie token-level and feature-level fusion.

For token-level fusion, features output from encoders are transformed into tokens and concatenated with text tokens before being sent into LLMs. A common and feasible solution is to leverage a group of learnable query tokens to extract information in a query-based manner~\cite{carion2020end}, which first has been implemented in BLIP-2~\cite{li2023blip}, and subsequently inherited by a variety of work~\cite{x-llm, instructblip, video-llama}. Such Q-Former-style approaches compress visual tokens into a smaller number of representation vectors.
In contrast, some methods simply use a MLP-based interface to bridge the modality gap~\cite{llava, pmc-vqa, pandagpt, detgpt}. 
For example, LLaVA series adopts one/two linear MLP~\cite{llava, liu2023improved} to project visual tokens and align the feature dimension with word embeddings.

On a related note, MM1~\cite{mckinzie2024mm1} has ablated on design choices on the connector and found that for token-level fusion, the type of modality adapter is far less important than the number of visual tokens and input resolution. 
Nevertheless, Zeng~\etal~\cite{zeng2023matters} compare the performance of token and feature-level fusion, and empirically reveal that the token-level fusion variant performs better in terms of VQA benchmarks. Regarding the performance gap, the authors suggest that cross-attention models might require a more complicated hyper-parameter searching process to achieve comparable performance. 

As another line, feature-level fusion inserts extra modules that enable deep interaction and fusion between text features and visual features. For example, Flamingo~\cite{alayrac2022flamingo} inserts extra cross-attention layers between frozen Transformer layers of LLMs, thereby augmenting language features with external visual cues. Similarly, CogVLM~\cite{wang2023cogvlm} plugs in a visual expert module in each Transformer layer to enable dual interaction and fusion between vision and language features. For better performance, the QKV weight matrix of the introduced module is initialized from the pre-trained LLM.
Similarly, LLaMA-Adapter~\cite{llama-adapter} introduces learnable prompts into Transformer layers. These prompts are first embedded with visual knowledge and then concatenated with text features as prefixes.

In terms of parameter size, learnable interfaces generally comprise a small portion compared with encoders and LLMs. Take Qwen-VL~\cite{bai2023qwen} as an example, the parameter size of the Q-Former is about 0.08B, accounting for less than 1\% of the whole parameters, while the encoder and the LLM account for about 19.8\% (1.9B) and 80.2\% (7.7B), respectively.

\noindent \textbf{Expert Model.}
Apart from the learnable interface, using expert models, such as an image captioning model, is also a feasible way to bridge the modality gap~\cite{yin2023woodpecker,guo2023images,CAT,chatcaptioner}.
The basic idea is to convert multimodal inputs into languages without training.
In this way, LLMs can understand multimodality by the converted languages.
For example, VideoChat-Text~\cite{videochat} uses pre-trained vision models to extract visual information such as actions and enriches the descriptions using a speech recognition model. 
Though using expert models is straightforward, it may not be as flexible as adopting a learnable interface. The conversion of foreign modalities into text would cause information loss. For example, transforming videos into textual descriptions distorts spatial-temporal relationships~\cite{videochat}.

\section{Training Strategy and Data}
\label{sec:mllm_train}
A full-fledged MLLM undergoes three stages of training, \ie pre-training, instruction-tuning, and alignment tuning. Each phase of training requires different types of data and fulfills different objectives. In this section, we discuss training objectives, as well as data collection and characteristics for each training stage.

\subsection{Pre-training}
\label{sec:train_pre_train}

\subsubsection{Training Detail}
As the first training stage, pre-training mainly aims to align different modalities and learn multimodal world knowledge.
Pre-training stage generally entails large-scale text-paired data, \eg caption data. Typically, the caption pairs describe images/audio/videos in natural language sentences.

Here, we consider a common scenario where MLLMs are trained to align vision with text. As illustrated in~\cref{pretrain_prompt}, given an image, the model is trained to predict autoregressively the caption of the image, following a standard cross-entropy loss.
A common approach for pre-training is to keep pre-trained modules (\eg visual encoders and LLMs) frozen and train a learnable interface~\cite{detgpt, llava, llava-med}. The idea is to align different modalities without losing pre-trained knowledge. Some methods~\cite{mplug-owl,visionllm,bai2023qwen} also unfreeze more modules (\eg visual encoder) to enable more trainable parameters for alignment. 
It should be noted that the training scheme is closely related to the data quality. 
For short and noisy caption data, a lower resolution (\eg 224) can be adopted to speed up the training process, while for longer and cleaner data, it is better to utilize higher resolutions (\eg 448 or higher) to mitigate hallucinations. 
Besides, ShareGPT4V~\cite{chen2023sharegpt4v} finds that with high-quality caption data in the pretraining stage, unlocking the vision encode promotes better alignment. 

\begin{table}[!t]
\begin{tcolorbox}

Input: \textcolor[rgb]{0,0,0.8}{<image>}\ \par

Response: \textcolor[rgb]{0.8,0,0}{\{caption\}}

\end{tcolorbox}
\caption{A simplified template to structure the caption data. \textcolor[rgb]{0,0,0.8}{\{<image>\}} is the placeholder for the visual tokens, and \textcolor[rgb]{0.8,0,0}{\{caption\}} is the caption for the image. Note that only the part marked in \textcolor[rgb]{0.8,0,0}{red} is used for loss calculation.}
\label{pretrain_prompt}
\end{table}

\subsubsection{Data}
Pretraining data mainly serve two purposes, \ie (1) aligning different modalities and (2) providing world knowledge. 
The pretraining corpora can be divided into coarse-grained and fine-grained data according to granularities, which we will introduce sequentially. We summarize commonly used pretraining datasets in~\cref{tab:data-pretrain}.

Coarse-grained caption data share some typical traits in common: (1) The data volume is large since samples are generally sourced from the internet. (2) Because of the web-scrawled nature, the captions are usually short and noisy since they originate from the alt-text of the web images. 
These data can be cleaned and filtered via automatic tools, for example, using CLIP~\cite{clip} model to filter out image-text pairs whose similarities are lower than a pre-defined threshold.
In what follows, we introduce some representative coarse-grained datasets.

\noindent \textbf{CC.} 
CC-3M~\cite{CC} is a web-scale caption dataset of 3.3M image-caption pairs, where the raw descriptions are derived from alt-text associated with images. 
The authors design a complicated pipeline to clean data:
(1) For images, those with inappropriate content or aspect ratio are filtered. (2) For text, NLP tools are used to obtain text annotations, with samples filtered according to the designed heuristics. (3) For image-text pairs, images are assigned labels via classifiers. If text annotations do not overlap with image labels, the corresponding samples are dropped.

CC-12M~\cite{CC-12m} is a following work of CC-3M and contains 12.4M image-caption pairs. Compared with the previous work, CC-12M relaxes and simplifies the data-collection pipeline, thus collecting more data.

\noindent \textbf{SBU Captions~\cite{SBU-Captions}.} 
It is a captioned photo dataset containing 1M image-text pairs, with images and descriptions sourced from Flickr. 
Specifically, an initial set of images is acquired by querying the Flickr website with a large number of query terms. The descriptions attached to the images thus serve as captions. Then, to ensure that descriptions are relevant to the images, the retained images fulfill these requirements: (1) Descriptions of the images are of satisfactory length, decided by observation. (2) Descriptions of the images contain at least 2 words in the predefined term lists and a propositional word (\eg ``on'', ``under'') that generally suggests spatial relationships.

\noindent \textbf{LAION.}
This series are large web-scale datasets, with images scrawled from the internet and associated alt-text as captions. To filter the image-text pairs, the following steps are performed: (1) Text with short lengths or images with too small or too big sizes are dropped. (2) Image deduplication based on URL. (3) Extract CLIP~\cite{clip} embeddings for images and text, and use the embeddings to drop possibly illegal content and image-text pairs with low cosine similarity between embeddings.
Here we offer a brief summary of some typical variants:
\begin{itemize}[leftmargin=*]
    \item LAION-5B~\cite{laion-5b}: It is a research-purpose dataset of 5.85B image-text pairs. The dataset is multilingual with a 2B English subset.
    \item LAION-COCO~\cite{laion-coco}: It contains 600M images extracted from the English subset of LAION-5B. The captions are synthetic, using BLIP~\cite{li2022blip} to generate various image captions and using CLIP~\cite{clip} to pick the best fit for the image.
\end{itemize}

\begin{table}[!t]
\centering
\caption{Common datasets used for pre-training.}
\label{tab:data-pretrain}
\begin{tabular}{lcc}
\toprule
\textbf{Dataset}               & \textbf{Samples}   & \textbf{Date}             \\
\midrule
\multicolumn{3}{l}{\textbf{Coarse-grained Image-Text}}                             \\
\midrule
CC-3M~\cite{CC}                          & 3.3M                  & 2018                      \\
CC-12M~\cite{CC-12m}                         & 12.4M                 & 2020                      \\
SBU Captions~\cite{SBU-Captions}                   & 1M                    & 2011                      \\
LAION-5B~\cite{laion-5b}                       & 5.9B                  & Mar-2022                  \\
LAION-2B~\cite{laion-5b}                       & 2.3B                  & Mar-2022                  \\
LAION-COCO~\cite{laion-coco}                     & 600M                  & Sep-2022                  \\
COYO-700M~\cite{kakaobrain2022coyo-700m}                      & 747M                  & Aug-2022                  \\
\midrule
\multicolumn{3}{l}{\textbf{Fine-grained Image-Text}}                               \\
\midrule
ShareGPT4V-PT~\cite{chen2023sharegpt4v}    & 1.2M                & Nov-2023      \\
LVIS-Instruct4V~\cite{wang2023see}                & 111K                  & Nov-2023                  \\
ALLaVA~\cite{chen2024allava}                         & 709K                  & Feb-2024                  \\
\midrule
\multicolumn{3}{l}{\textbf{Video-Text}}                                            \\
\midrule
MSR-VTT~\cite{msr-vtt}                        & 200K                  & 2016                      \\
\midrule
\multicolumn{3}{l}{\textbf{Audio-Text}}                                            \\
\midrule
WavCaps~\cite{wavcaps}                        & 24K                   & Mar-2023            \\
\bottomrule
\end{tabular}
\end{table}

\noindent \textbf{COYO-700M~\cite{kakaobrain2022coyo-700m}.}
It contains 747M image-text pairs, which are extracted from CommonCrawl.
For data filtering, the authors design the following strategies:
(1) For images, those with inappropriate size, content, format, or aspect ratio are filtered. Moreover, the images are filtered based on the pHash value to remove images overlapped with public datasets such as ImageNet and MS-COCO. (2) For text, only English text with satisfactory length, noun forms, and appropriate words are saved. Whitespace before and after the sentence will be removed, and consecutive whitespace characters will be replaced with a single whitespace. Moreover, text appearing more than 10 times (\eg ``image for'') will be dropped. (3) For image-text pairs, duplicated samples are removed based on (image pHash, text) tuple.

Recently, more works~\cite{chen2023sharegpt4v,wang2023see,chen2024allava} have explored generating high-quality fine-grained data through prompting strong MLLMs (\eg GPT-4V). Compared with coarse-grained data, these data generally contain longer and more accurate descriptions of the images, thus enabling finer-grained alignment between image and text modalities. However, since this approach generally requires calling commercial-use MLLMs, the cost is higher, and the data volume is relatively smaller.  Notably, ShareGPT4V~\cite{chen2023sharegpt4v} strikes a balance by first training a captioner with GPT-4V-generated 100K data, then scaling up the data volume to 1.2M using the pre-trained captioner.

\subsection{Instruction-tuning}
\label{sec:train_inst_tune}

\begin{figure}[t]
    \centering
    \includegraphics[width=0.985\linewidth]{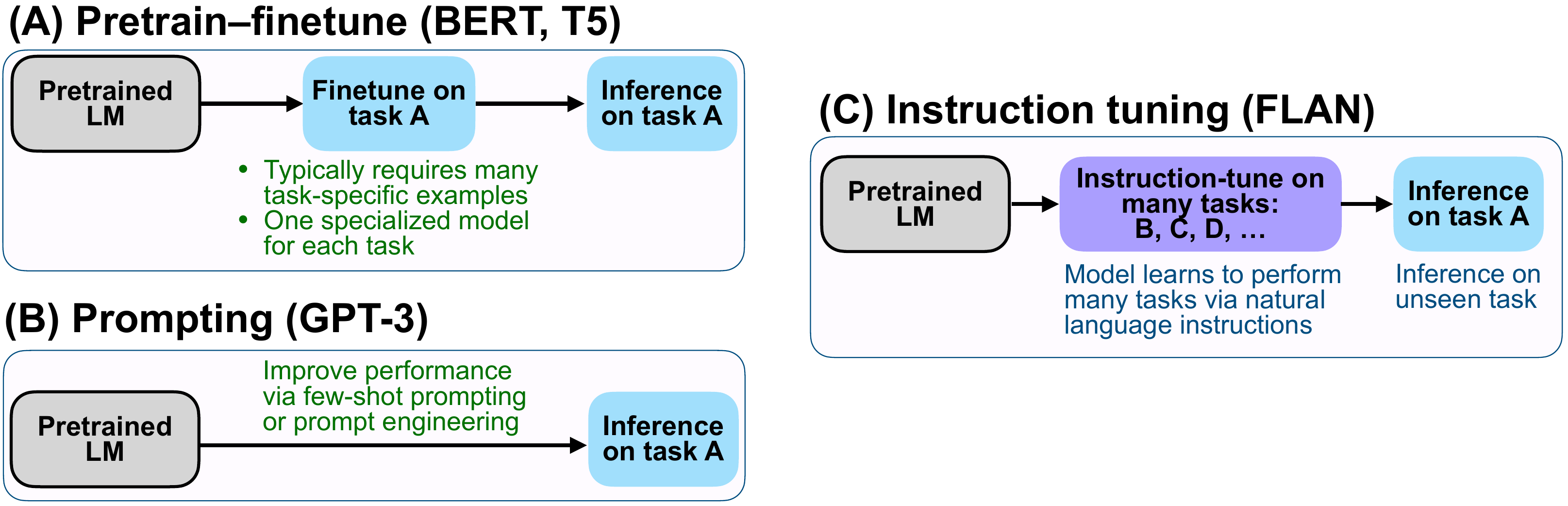}
    \caption{Comparison of three typical learning paradigms. The image is from~\cite{wei2021finetuned}.}
    \label{fig:compare_learn}
\end{figure}

\subsubsection{Introduction}
Instruction refers to the description of tasks.
Intuitively, instruction tuning aims to teach models to better understand the instructions from users and fulfill the demanded tasks. 
Tuning in this way, LLMs can generalize to unseen tasks by following new instructions, thus boosting zero-shot performance. 
This simple yet effective idea has sparked the success of subsequent NLP works, such as ChatGPT~\cite{chatgpt}, InstructGPT~\cite{ouyang2022training}, FLAN~\cite{chung2022scaling, wei2021finetuned}, and OPT-IML~\cite{iyer2022opt}. 

The comparisons between instruction tuning and related typical learning paradigms are illustrated in \cref{fig:compare_learn}. 
The supervised fine-tuning approach usually requires a large amount of task-specific data to train a task-specific model.
The prompting approach reduces the reliance on large-scale data and can fulfill a specialized task via prompt engineering.
In such a case, though the few-shot performance has been improved, the zero-shot performance is still quite average~\cite{brown2020language}.
Differently, instruction tuning learns how to generalize to unseen tasks rather than fitting specific tasks like the two counterparts.
Moreover, instruction tuning is highly related to multi-task prompting~\cite{sanh2021multitask}.

In this section, we delineate the format of instruction samples, the training objectives, typical ways to gather instruction data, and corresponding commonly used datasets.

\subsubsection{Training Detail}
\label{sec:train_inst_tune_detail}

\begin{table}[!b]
\begin{tcolorbox}
Below is an instruction that describes a task. Write a response that appropriately completes the request \par\medskip

Instruction: \textcolor[rgb]{0,0,0.8}{<instruction>}

Input: \textcolor[rgb]{0,0,0.8}{\{<image>, <text>\}}\ 

Response: \textcolor[rgb]{0.8,0,0}{<output>}

\end{tcolorbox}
\caption{A simplified template to structure the multimodal instruction data. \textcolor[rgb]{0,0,0.8}{<instruction>} is a textual description of the task. \textcolor[rgb]{0,0,0.8}{\{<image>, <text>\}} and \textcolor[rgb]{0.8,0,0}{<output>} are input and output from the data sample. Note that \textcolor[rgb]{0,0,0.8}{<text>} in the input may be missed for some datasets, such as image caption datasets merely have \textcolor[rgb]{0,0,0.8}{<image>}. The example is adapted from~\cite{multimodal-gpt}.}
\label{vision_prompt}
\end{table}

A multimodal instruction sample often includes an optional instruction and an input-output pair. The instruction is typically a natural language sentence describing the task, such as, ``\textit{Describe the image in detail.}'' The input can be an image-text pair like the VQA task~\cite{antol2015vqa} or only an image like the image caption task~\cite{karpathy2015deep}. The output is the answer to the instruction conditioned on the input.
The instruction template is flexible and subject to manual designs~\cite{videochat, multimodal-gpt, llava}, as exemplified in \cref{vision_prompt}. Note that the instruction template can also be generalized to the case of multi-round conversations~\cite{multimodal-gpt, pmc-vqa, llava, pandagpt}.

Formally, a multimodal instruction sample can be denoted in a triplet form, \ie $(\mathcal{I}, \mathcal{M}, \mathcal{R})$, where $\mathcal{I}, \mathcal{M}, \mathcal{R}$ represent the instruction, the multimodal input, and the ground truth response, respectively. The MLLM predicts an answer given the instruction and the multimodal input:
\begin{equation}
    \mathcal{A} = f(\mathcal{I}, \mathcal{M}; \theta)
\end{equation}
Here, $\mathcal{A}$ denotes the predicted answer, and $\theta$ are the parameters of the model. 
The training objective is typically the original auto-regressive objective used to train LLMs~\cite{llava, pmc-vqa, pandagpt, lavin}, based on which the MLLM is encouraged to predict the next token of the response. The objective can be expressed as:
\begin{equation}
\mathcal{L}(\theta) = -\sum_{i=1}^{N} \log p(\mathcal{R}_i|\mathcal{I}, \mathcal{R}_{<i}; \theta)
\end{equation}
where $N$ is the length of the ground-truth response. 

\subsubsection{Data Collection}
\label{sec:train_inst_tune_collect}
Since instruction data are more flexible in formats and varied in task formulations, it is usually trickier and more costly to collect data samples. In this section, we summarize three typical ways to harvest instruction data at scale, \ie data adaptation, self-instruction, and data mixture.

\begin{table*}[!htbp]
\begin{tcolorbox}
\centering
\small
\begin{itemize}[leftmargin=2.5mm]
\setlength{\itemsep}{1pt}
\item <Image> \{Question\}
\item <Image> Question: \{Question\}
\item <Image> \{Question\} A short answer to the question is
\item <Image> Q: \{Question\} A:
\item <Image> Question: \{Question\} Short answer:
\item <Image> Given the image, answer the following question with no more than three words. \{Question\}
\item <Image> Based on the image, respond to this question with a short answer: \{Question\}. Answer:
\item <Image> Use the provided image to answer the question: \{Question\} Provide your answer as short as possible:
\item <Image> What is the answer to the following question? "\{Question\}"
\item <Image> The question "\{Question\}" can be answered using the image. A short answer is
\end{itemize}
\end{tcolorbox}
\caption{Instruction templates for VQA datasets, cited from~\cite{instructblip}. <Image> and \{Question\} are the image and the question in the original VQA datasets, respectively.}
\label{tab:vqa_instructions}
\end{table*}

\noindent \textbf{Data Adaptation.}
Task-specific datasets are rich sources of high-quality data. Hence, abundant works~\cite{instructblip, visionllm, x-llm, multiinstruct, llama-adapter, llama-adapter-v2, chatbridge, lavin} have utilized existing high-quality datasets to construct instruction-formatted datasets. Take the transformation of VQA datasets for an example, the original sample is an input-out pair where the input comprises an image and a natural language question, and the output is the textual answer to the question conditioned on the image. The input-output pairs of these datasets could naturally comprise the multimodal input and response of the instruction sample (see \S\ref{sec:train_inst_tune_detail}). The instructions, \ie the descriptions of the tasks, can either derive from manual design or from semi-automatic generation aided by GPT. Specifically, some works~\cite{instructblip, multiinstruct, minigpt-4, x-llm, llava-med, m3it} hand-craft a pool of candidate instructions and sample one of them during training. We offer an example of instruction templates for the VQA datasets as shown in~\cref{tab:vqa_instructions}. The other works manually design some seed instructions and use these to prompt GPT to generate more~\cite{visionllm, videochat, multimodal-gpt}. 

Note that since the answers of existing VQA and caption datasets are usually concise, directly using these datasets for instruction tuning may limit the output length of MLLMs. There are two common strategies to tackle this problem. 
The first one is to specify explicitly in instructions. 
For example, ChatBridge~\cite{chatbridge} explicitly declares \textit{short} and \textit{brief} for short-answer data, as well as \textit{a sentence} and \textit{single sentence} for conventional coarse-grained caption data.
The second one is to extend the length of existing answers~\cite{m3it}.
For example, M$^3$IT~\cite{m3it} proposes to rephrase the original answer by prompting ChatGPT with the original question, answer, and contextual information of the image (\eg caption and OCR).

\begin{table*}[!htb]
\centering
\caption{A summary of popular datasets generated by self-instruction. For input/output modalities, I: Image, T: Text, V: Video, A: Audio. For data composition, M-T and S-T denote multi-turn and single-turn, respectively.}
\label{tab:IT-data}
\begin{tabular}{lcccc}
\toprule
\textbf{Dataset} & \textbf{Sample} & \textbf{Modality}     & \textbf{Source} & \textbf{Composition}                     \\
\midrule
LLaVA-Instruct   & 158K              & I + T $\rightarrow$ T & MS-COCO         & 23K caption + 58K M-T QA + 77K reasoning \\
LVIS-Instruct & 220K & I + T $\rightarrow$ T & LVIS         & 110K caption + 110K M-T QA             \\
ALLaVA        & 1.4M & I + T $\rightarrow$ T & VFlan, LAION & 709K caption + 709K S-T QA             \\
\midrule
Video-ChatGPT & 100K & V + T $\rightarrow$ T & ActivityNet  & 7K description + 4K M-T QA             \\
VideoChat     & 11K  & V+T $\rightarrow$ T   & WebVid       & description + summarization + creation \\
\midrule
Clotho-Detail & 3.9K & A + T $\rightarrow$ T & Clotho       & caption                                \\
\bottomrule
\end{tabular}
\end{table*}

\noindent \textbf{Self-Instruction.}
Although existing multi-task datasets can contribute a rich source of data, they usually do not meet human needs well in real-world scenarios, such as multiple rounds of conversations.
To tackle this issue, some works collect samples through self-instruction~\cite{wang2022self}, which utilizes LLMs to generate textual instruction-following data using a few hand-annotated samples. Specifically, some instruction-following samples are hand-crafted as demonstrations, after which ChatGPT/GPT-4 is prompted to generate more instruction samples with the demonstrations as guidance. LLaVA~\cite{llava} extends the approach to the multimodal field by translating images into text of captions and bounding boxes, and prompting text-only GPT-4 to generate new data with the guidance of requirements and demonstrations. 
In this way, a multimodal instruction dataset is constructed, called LLaVA-Instruct-150k. Following this idea, subsequent works such as MiniGPT-4~\cite{minigpt-4}, ChatBridge~\cite{chatbridge}, GPT4Tools~\cite{gpt4tools}, and DetGPT~\cite{detgpt} develop different datasets catering for different needs.
Recently, with the release of the more powerful multimodal model GPT-4V, many works have adopted GPT-4V to generate data of higher quality, as exemplified by LVIS-Instruct4V~\cite{wang2023see} and ALLaVA~\cite{chen2024allava}. We summarize the popular datasets generated through self-instruction in~\cref{tab:IT-data}.

\noindent \textbf{Data Mixture.}
Apart from the multimodal instruction data, language-only user-assistant conversation data can also be used to improve conversational proficiencies and instruction-following abilities~\cite{llama-adapter-v2, mplug-owl, multimodal-gpt, lavin}. 
LaVIN~\cite{lavin} directly constructs a minibatch by randomly sampling from both language-only and multimodal data. 
MultiInstruct~\cite{multiinstruct} probes different strategies for training with a fusion of single modal and multimodal data, including mixed instruction tuning (combine both types of data and randomly shuffle) and sequential instruction tuning (text data followed by multimodal data).

\subsubsection{Data Quality}
Recent research has revealed that the data quality of instruction-tuning samples is no less important than quantity.
Lynx~\cite{zeng2023matters} finds that models pre-trained on large-scale but noisy image-text pairs do not perform as well as models pre-trained with smaller but cleaner datasets.
Similarly, Wei~\etal~\cite{wei2023instructiongpt} finds that less instruction-tuning data with higher quality can achieve better performance. For data filtering, the work proposes some metrics to evaluate data quality and, correspondingly, a method to automatically filter out inferior vision-language data.
Here we discuss two important aspects regarding data quality.  

\noindent \textbf{Prompt Diversity.}
The diversity of instructions has been found to be critical for model performance. Lynx~\cite{zeng2023matters} empirically verifies that diverse prompts help improve model performance and generalization ability. 

\noindent \textbf{Task Coverage.}
In terms of tasks involved in training data, Du~\etal~\cite{du2023makes} perform an empirical study and find that the visual reasoning task is superior to captioning and QA tasks for boosting model performance. Moreover, the study suggests that enhancing the complexity of instructions might be more beneficial than increasing task diversity and incorporating fine-grained spatial annotations.

\subsection{Alignment tuning}
\label{sec:train_align_tune}

\subsubsection{Introduction}
Alignment tuning is more often used in scenarios where models need to be aligned with specific human preferences, \eg response with fewer hallucinations (see \S \ref{sec:hallu}).
Currently, Reinforcement Learning with Human Feedback (RLHF) and Direct Preference Optimization (DPO) are two main techniques for alignment tuning. In this section, we introduce the main ideas of the two techniques in sequence and offer some examples of how they are utilized in addressing practical problems, and finally, give a compilation of the related datasets.

\subsubsection{Training Detail}

\noindent \textbf{RLHF~\cite{ziegler2019fine, stiennon2020learning}.}
This technique aims to utilize reinforcement learning algorithms to align LLMs with human preferences, with human annotations as supervision in the training loop. As exemplified in InstructGPT~\cite{ouyang2022training}, RLHF incorporates three key steps:

\begin{enumerate}[leftmargin=*]
    \item \textbf{Supervised fine-tuning.} This step aims to fine-tune a pre-trained model to present the preliminary desired output behavior. The fine-tuned model in the RLHF setting is called a \textit{policy model}. Note that this step might be skipped since the supervised policy model $\pi^{\text{SFT}}$ can be initialized from an instruction-tuned model (see \S \ref{sec:train_inst_tune}). 
    \item \textbf{Reward modeling.} A \textit{reward model} is trained using preference pairs in this step. Given a multimodal prompt (\eg image and text) $x$ and a response pair $(y_w, y_l)$, the reward model $r_\theta$ learns to give a higher reward to the preferred response $y_w$, and vice versa for $y_l$, according to the following objective:
    \begin{equation}
        \mathcal{L}(\theta)=-\mathbb{E}_{(x,y_w,y_l)\sim \mathcal{D}} \left[ \log (\sigma(r_\theta (x, y_w) - r_\theta(x, y_l) \right]
    \end{equation}
    where $\mathcal{D}=\{(x,y_w,y_l)\}$ is the comparison dataset labeled by human annotators. In practice, the reward model $r_\theta$ shares a similar structure with the policy model.
    \item \textbf{Reinforcement learning.} In this step, the Proximal Policy Optimization (PPO) algorithm is adopted to optimize the RL policy model $\pi^{\text{RL}}_\phi$. A per-token KL penalty is often added to the training objective to avoid deviating too far from the original policy~\cite{ouyang2022training}, resulting in the objective:
     \begin{equation}
     \begin{split}
        \mathcal{L}(\phi) &= -\mathbb{E}_{x\sim \mathcal{D}, y\sim \pi^{RL}_{\phi}(y|x)} \Big[ r_\theta(x,y)  \\
        &- \beta\cdot \mathbb{D}_{KL}\Big(\pi^{RL}_{\phi}(y|x) || \pi^{REF}(y|x)\Big) \Big]
    \end{split}
    \end{equation}
    where $\beta$ is the coefficient for the KL penalty term. Typically, both the RL policy $\pi^{\text{RL}}_\phi$ and the reference model $\pi^{\text{REF}}$ are initialized from the supervised model $\pi^{\text{SFT}}$. The obtained RL policy model is expected to align with human preferences through this tuning process.    
\end{enumerate}

Researchers have explored using the RLHF techniques for better multimodal alignment. For example, LLaVA-RLHF~\cite{sun2023aligning} collects human preference data and tunes a model with fewer hallucinations based on LLaVA~\cite{llava}.

\noindent \textbf{DPO~\cite{rafailov2023direct}.}
It learns from human preference labels utilizing a simple binary classification loss. Compared with the PPO-based RLHF algorithm, DPO is exempt from learning an explicit reward model, thus simplifying the whole pipeline to two steps, \ie human preference data collection and preference learning. The learning objective is as follows:
\begin{equation}
\begin{split}
    \mathcal{L}(\phi) &= -\mathbb{E}_{(x,y_w,y_l)\sim \mathcal{D}} \Big[ \log \sigma\Big( \beta \log \frac{\pi_\phi^{\text{RL}}(y_w|x)}{\pi^{\text{REF}}(y_w|x)} \\ 
    &- \beta \log \frac{\pi_\phi^{\text{RL}}(y_l|x)}{\pi^{\text{REF}}(y_l|x)} \Big)\Big]
\end{split}
\end{equation}
RLHF-V~\cite{yu2023rlhf} collects fine-grained (segment-level) preference data pairs by correcting hallucinations in the model response and uses the obtained data to perform dense DPO.
Silkie\cite{li2023silkie} instead collects preference data via prompting GPT-4V and distills the preference supervision into an instruction-tuned model through DPO.

\subsubsection{Data}
The gist of data collection for alignment-tuning is to collect feedback for model responses, \ie to decide which response is better. 
It is generally more expensive to collect such data, and the amount of data used for this phase is typically even less than that used in previous stages. 
In this part, we introduce some datasets and summarize them in~\cref{tab:align-tune-data}.

\noindent \textbf{LLaVA-RLHF~\cite{sun2023aligning}.}
It contains 10K preference pairs collected from human feedback in terms of honesty and helpfulness. The dataset mainly serves to reduce hallucinations in model responses.

\noindent \textbf{RLHF-V~\cite{yu2023rlhf}.}
It has 5.7K fine-grained human feedback data collected by segment-level hallucination corrections.

\noindent \textbf{VLFeedback~\cite{li2023silkie}.}
It utilizes AI to provide feedback on model responses. The dataset contains more than 380K comparison pairs scored by GPT-4V in terms of helpfulness, faithfulness, and ethical concerns.

\begin{table}[!t]
\centering
\caption{A summary of datasets for alignment-tuning. For input/output modalities, I: Image, T: Text.}
\label{tab:align-tune-data}
\begin{tabular}{lccc}
\toprule
\textbf{Dataset} & \textbf{Sample} & \textbf{Modality}     & \textbf{Source} \\
\midrule
LLaVA-RLHF~\cite{sun2023aligning}       & 10K               & I + T $\rightarrow$ T & Human           \\
RLHF-V~\cite{yu2023rlhf}           & 5.7K              & I + T $\rightarrow$ T & Human           \\
VLFeedback~\cite{li2023silkie}       & 380K              & I + T $\rightarrow$ T & GPT-4V          \\
\bottomrule
\end{tabular}
\end{table}

\section{Evaluation}
\label{sec:mllm_eval}
Evaluation is an essential part of developing MLLMs since it provides feedback for model optimization and helps to compare the performance of different models. Compared with evaluation methods of traditional multimodal models, the evaluation of MLLMs exhibits several new traits: (1) Since MLLMs are generally versatile, it is important to evaluate MLLMs comprehensively. (2) MLLMs exhibit many emergent capabilities that require special attention (\eg OCR-free math reasoning) and thus require new evaluation schemes.
The evaluation of MLLMs can be broadly categorized into two types according to the question genres, including closed-set and open-set. 

\subsection{Closed-set}
Closed-set questions refer to a type of question where the possible answer options are predefined and limited to a finite set. 
The evaluation is usually performed on task-specific datasets. 
In this case, the responses can be naturally judged by benchmark metrics~\cite{instructblip, llava, x-llm, multiinstruct, llama-adapter, llama-adapter-v2, chatbridge, lavin}. 
For example, InstructBLIP~\cite{instructblip} reports the accuracy on ScienceQA~\cite{scienceqa}, as well as the CIDEr score~\cite{vedantam2015cider} on NoCaps~\cite{agrawal2019nocaps} and Flickr30K~\cite{young2014image}. 
The evaluation settings are typically zero-shot~\cite{instructblip, multiinstruct, chatbridge, m3it} or finetuning~\cite{instructblip, llava, x-llm, llama-adapter, llama-adapter-v2, lavin, llava-med, m3it}. 
The first setting often selects a wide range of datasets covering different general tasks and splits them into held-in and held-out datasets. 
After tuning on the former, zero-shot performance is evaluated on the latter with unseen datasets or even unseen tasks. 
In contrast, the second setting is often observed in the evaluation of domain-specific tasks. 
For example, LLaVA~\cite{llava} and LLaMA-Adapter~\cite{llama-adapter} report finetuned performance on ScienceQA~\cite{scienceqa}. 
LLaVA-Med~\cite{llava-med} reports results on biomedical VQA~\cite{he2020pathvqa, lau2018dataset, liu2021slake}.

The above evaluation methods are usually limited to a small range of selected tasks or datasets, lacking a comprehensive quantitative comparison. 
To this end, some efforts have endeavored to develop new benchmarks specially designed for MLLMs~\cite{mme,mmbench,mm-vet,seed-bench,mathvista,mmmu,hallusionbench}. 
For example, Fu \etal~\cite{mme} construct a comprehensive evaluation benchmark MME that includes a total of 14 perception and cognition tasks.
All instruction-answer pairs in MME are manually designed to avoid data leakage.
MMBench~\cite{mmbench} is a benchmark specifically designed for evaluating multiple dimensions of model capabilities, using ChatGPT to match open responses with pre-defined choices.
Video-ChatGPT~\cite{video-chatgpt} and Video-Bench~\cite{ning2023video} focus on video domains and propose specialized benchmarks as well as evaluation tools for assessment.
There are also evaluation strategies designed to evaluate a specific aspect of the model~\cite{multiinstruct}, as exemplified by POPE~\cite{li2023evaluating} for assessment of hallucination degree.

\subsection{Open-set}
In contrast to the closed-set questions, the responses to open-set questions can be more flexible, where MLLMs usually play a chatbot role.
Because the content of the chat can be arbitrary, it would be trickier to judge than the closed-ended output.
The criterion can be classified into manual scoring, GPT scoring, and case study. 
Manual scoring requires humans to assess the generated responses. 
This kind of approach often involves hand-crafted questions that are designed to assess specific dimensions. 
For example, mPLUG-Owl~\cite{mplug-owl} collects a visually related evaluation set to judge capabilities like natural image understanding, diagram, and flowchart understanding.  
Similarly, GPT4Tools~\cite{gpt4tools} builds two sets for the finetuning and zero-shot performance, respectively, and evaluates the responses in terms of thought, action, arguments, and the whole.

Since manual assessment is labor intensive, some researchers have explored rating with GPT, namely GPT scoring. 
This approach is often used to evaluate performance on multimodal dialogue. 
LLaVA~\cite{llava} proposes to score the responses via text-only GPT-4 in terms of different aspects, such as helpfulness and accuracy. 
Specifically, 30 images are sampled from the COCO~\cite{COCO} validation set, each associated with a short question, a detailed question, and a complex reasoning question via self-instruction on GPT-4. 
The answers generated by both the model and GPT-4 are sent to GPT-4 for comparison. 
Subsequent works follow this idea and prompt ChatGPT~\cite{mplug-owl} or GPT-4~\cite{x-llm, chatbridge, lavin, llava-med, m3it} to rate results~\cite{mplug-owl, x-llm, chatbridge, lavin, llava-med} or judge which one is better~\cite{llama-adapter-v2}.

A main issue of applying text-only GPT-4 as an evaluator is that the judge is only based on image-related text content, such as captions or bounding box coordinates, without accessing the image~\cite{llava-med}. 
Thus, it may be questionable to set GPT-4 as the performance upper bound in this case. With the release of the vision interface of GPT, some works~\cite{yin2023woodpecker, li2024red} exploit a more advanced GPT-4V model to assess the performance of MLLMs. For example, Woodpecker~\cite{yin2023woodpecker} adopts GPT-4V to judge the response quality of model answers based on the image. The evaluation is expected to be more accurate than using text-only GPT-4 since GPT-4V has direct access to the image.

A supplementary approach is to compare the different capabilities of MLLMs through case studies.
For instance, some studies evaluate two typical advanced commercial-use models, GPT-4V and Gemini. 
Yang~\etal~\cite{yang2023dawn} perform in-depth qualitative analysis on GPT-4V by crafting a series of samples across various domains and tasks, spanning from preliminary skills, such as caption and object counting, to complex tasks that require world knowledge and reasoning, such as joke understanding and indoor navigation as an embodied agent. 
Wen~\etal~\cite{wen2023road} make a more focused evaluation of GPT-4V by designing samples targeting automatic driving scenarios.
Fu~\etal~\cite{fu2023challenger} carry out a comprehensive evaluation on Gemini-Pro by comparing the model against GPT-4V. The results suggest that GPT-4V and Gemini exhibit comparable visual reasoning abilities in spite of different response styles.

\section{Extensions}
\label{sec:extensions}

Recent studies have made significant strides in extending the capabilities of MLLMs, spanning from more potent foundational abilities to broader coverage of scenarios. We trace the principal development of MLLMs in this regard.

\noindent \textbf{Granularity Support.}
To facilitate better interaction between agents and users, researchers have developed MLLMs with finer support of granularities in terms of model inputs and outputs.
On the input side, models that support finer control from user prompts are developed progressively, evolving from image to region~\cite{chen2023shikra,zhang2023gpt4roi,xuan2023pink} and even pixels~\cite{yuan2023osprey,rasheed2023glamm,you2023ferret}.
Specifically, Shikra~\cite{chen2023shikra} supports region-level input and understanding. Users may interact with the assistant more flexibly by referring to specific regions, which are represented in bounding boxes of natural language forms.
Ferret~\cite{you2023ferret} takes a step further and supports more flexible referring by devising a hybrid representation scheme. The model supports different forms of prompts, including point, box, and sketch. 
Similarly, Osprey~\cite{yuan2023osprey} supports point input by utilizing a segmentation model~\cite{sam}. Aided by the exceptional capabilities of the pre-trained segmentation model, Osprey enables specifying a single entity or part of it with a single click.
On the output side, grounding capabilities are improved in line with the development of input support. 
Shikra~\cite{chen2023shikra} supports response grounded in the image with box annotations, resulting in higher precision and finer referring experience.
LISA~\cite{lai2023lisa} further supports mask-level understanding and reasoning, which makes pixel-level grounding possible. 

\noindent \textbf{Modality Support.}
Increased support for modalities is a tendency for MLLM studies.
On the one hand, researchers have explored adapting MLLMs to support the input of more multimodal content, such as 3D point cloud~\cite{xu2023pointllm,chen2023ll3da,huang2023embodied,hong20233d}.
On the other hand, MLLMs are also extended to generate responses of more modalities, such as image~\cite{emu,zhan2024anygpt,wu2023next,aiello2023jointly}, audio~\cite{zhang2023speechgpt,rubenstein2023audiopalm,zhan2024anygpt,wu2023next}, and video~\cite{wu2023next,wang2024modaverse}. 
For example, NExT-GPT~\cite{wu2023next} proposes a framework that supports inputs and outputs of mixed modalities, specifically, combinations of text, image, audio, and video, with the help of diffusion models~\cite{ho2020denoising,rombach2022high} attached to the MLLM. The framework applies an encoder-decoder architecture and puts LLM as a pivot for understanding and reasoning.

\noindent \textbf{Language Support.}
Current models are predominantly unilingual, probably due to the fact that high-quality non-English training corpus is scarce. Some works have been devoted to developing multilingual models so that a broader range of users can be covered.
VisCPM~\cite{hu2023large} transfers model capabilities to the multilingual setting by designing a multi-stage training scheme. Specifically, the scheme takes English as a pivotal language, with abundant training corpus. Utilizing a pre-trained bilingual LLM, the multimodal capabilities are transferred to Chinese by adding some translated samples during instruction tuning. 
Taking a similar approach, Qwen-VL~\cite{bai2023qwen} is developed from the bilingual LLM Qwen~\cite{qwen-lm} and supports both Chinese and English. During pre-training, Chinese data is mixed into the training corpus to preserve the bilingual capabilities of the model, taking up 22.7\% of the whole data volume.

\noindent \textbf{Scenario/Task Extension.}
Apart from developing common general-purpose assistants, some studies have focused on more specific scenarios where practical conditions should be considered, while others extend MLLMs to downstream tasks with specific expertise.

A typical tendency is to adapt MLLMs to more specific real-life scenarios. MobileVLM~\cite{chu2023mobilevlm} explores developing small-size variants of MLLMs for resource-limited scenarios. Some designs and techniques are utilized for deployment on mobile devices, such as LLMs of smaller size and quantization techniques to speed up computation.
Other works develop agents that interact with real-world~\cite{gong2023mindagent,huang2023embodied,embodiedgpt}, \eg user-friendly assistants specially designed for Graphical User Interface (GUI), as exemplified by CogAgent~\cite{hong2023cogagent}, AppAgent~\cite{yang2023appagent}, and Mobile-Agent~\cite{wang2024mobile}. These assistants excel in planning and guiding through each step to fulfill a task specified by users, acting as helpful agents for human-machine interaction.
Another line is to augment MLLMs with specific skills for solving tasks in different domains, \eg document understanding~\cite{ye2023mplug,liu2024textmonkey,hu2024mplug,ye2023ureader} and medical domains~\cite{llava-med,pmc-vqa,moor2023med}.
For document understanding, mPLUG-DocOwl~\cite{ye2023mplug} utilizes various forms of document-level data for tuning, resulting in an enhanced model in OCR-free document understanding. TextMonkey~\cite{liu2024textmonkey} incorporates multiple tasks related to document understanding to improve model performance. Apart from conventional document image and scene text datasets, position-related tasks are added to reduce hallucinations and help models learn to ground responses in the visual information.
MLLMs can also be extended to medical domains by instilling knowledge of the medical domain. For example, LLaVA-Med~\cite{li2023llava} injects medical knowledge into vanilla LLaVA~\cite{llava} and develops an assistant specialized in medical image understanding and question answering.

\section{Multimodal Hallucination}
\label{sec:hallu}
Multimodal hallucination refers to the phenomenon of responses generated by MLLMs being inconsistent with the image content~\cite{yin2023woodpecker}. As a fundamental and important problem, the issue has received increased attention. In this section, we briefly introduce some related concepts and research development.

\subsection{Preliminaries}
\label{sec:hallu_prelim}
Current research on multimodal hallucinations can be further categorized into three types~\cite{zhai2023halle}:
\begin{enumerate}
    \item \textit{Existence Hallucination} is the most basic form, meaning that models incorrectly claim the existence of certain objects in the image.
    \item \textit{Attribute Hallucination} means describing the attributes of certain objects in a wrong way, \eg failure to identify a dog's color correctly. It is typically associated with existence hallucination since descriptions of the attributes should be grounded in objects present in the image.
    \item \textit{Relationship Hallucination} is a more complex type and is also based on the existence of objects. It refers to false descriptions of relationships between objects, such as relative positions and interactions.   
\end{enumerate}

In what follows, we first introduce some specific evaluation methods (\S \ref{sec:hallu_eval_method}), which are useful to gauge the performance of methods for mitigating hallucinations (\S \ref{sec:hallu_method}). Then, we will discuss in detail the current methods for reducing hallucinations, according to the main categories each method falls into.

\subsection{Evaluation Methods}
\label{sec:hallu_eval_method}

CHAIR~\cite{rohrbach2018object} is an early metric that evaluates hallucination levels in open-ended captions. The metric measures the proportion of sentences with hallucinated objects or hallucinated objects in all the objects mentioned.
In contrast, POPE~\cite{li2023evaluating} is a method that evaluates closed-set choices. Specifically, multiple prompts with binary choices are formulated, each querying if a specific object exists in the image. The method also covers more challenging settings to evaluate the robustness of MLLMs, with data statistics taken into consideration. The final evaluation uses a simple watchword mechanism, \ie by detecting keywords ``yes/no'', to convert open-ended responses into closed-set binary choices.
With a similar evaluation approach, MME~\cite{mme} provides a more comprehensive evaluation, covering aspects of existence, count, position and color, as exemplified in~\cite{yin2023woodpecker}.

Different from previous approaches that use matching mechanisms to detect and decide hallucinations, HaELM~\cite{wang2023evaluation} proposes using text-only LLMs as a judge to automatically decide whether MLLMs' captions are correct against reference captions. In light of the fact that text-only LLMs can only access limited image context and require reference annotations, Woodpecker~\cite{yin2023woodpecker} uses GPT-4V to directly assess model responses grounded in the image.
FaithScore~\cite{jing2023faithscore} is a more fine-grained metric based on a routine that breaks down descriptive sub-sentences and evaluates each sub-sentence separately.
Based on previous studies, AMBER~\cite{wang2023llm} is an LLM-free benchmark that encompasses both discriminative tasks and generative tasks and involves three types of possible hallucinations (see \S \ref{sec:hallu_prelim}).

\subsection{Mitigation Methods}
\label{sec:hallu_method}
According to high-level ideas, the current methods can be roughly divided into three categories: pre-correction, in-process-correction, and post-correction.

\noindent \textbf{Pre-correction.}
An intuitive and straightforward solution for hallucination is to collect specialized data (\eg negative data) and use the data for fine-tuning, thus resulting in models with fewer hallucinated responses. 

LRV-Instruction~\cite{liu2024mitigating} introduces a visual instruction tuning dataset. Apart from common positive instructions, the dataset incorporates delicately designed negative instructions at different semantic levels to encourage responses faithful to the image content.
LLaVA-RLHF~\cite{sun2023aligning} collects human-preference pairs and finetunes models with reinforcement learning techniques, leading to models more aligned with less hallucinated answers.

\noindent \textbf{In-process-correction.}
Another line is to make improvements in architectural design or feature representation. These works try to explore the reasons for hallucinations and design corresponding remedies to mitigate them in the generation process.

HallE-Switch~\cite{zhai2023halle} performs an empirical analysis of possible factors of object existence hallucinations and hypothesizes that existence hallucinations derive from objects not grounded by visual encoders, and they are actually inferred based on knowledge embedded in the LLM. Based on the assumption, a continuous controlling factor and corresponding training scheme are introduced to control the extent of imagination in model output during inference. 

VCD~\cite{leng2023mitigating} suggests that object hallucinations derive from two primary causes, \ie statistical bias in training corpus and strong language prior embedded in LLMs. The authors take notice of the phenomenon that when injecting noise into the image, MLLMs tend to lean towards language prior rather than the image content for response generation, leading to hallucinations. Correspondingly, this work designs an amplify-then-contrast decoding scheme to offset the false bias.

HACL~\cite{jiang2023hallucination} investigates the embedding space of vision and language. Based on the observation, a contrastive learning scheme is devised to pull paired cross-modal representation closer while pushing away non-hallucinated and hallucinated text representation.

\noindent \textbf{Post-correction.}
Different from previous paradigms, post-correction mitigates hallucinations in a post-remedy way and corrects hallucinations after output generation. 
Woodpecker~\cite{yin2023woodpecker} is a training-free general framework for hallucination correction. Specifically, the method incorporates expert models to supplement contextual information of the image and crafts a pipeline to correct hallucinations step by step. The method is interpretable in that intermediate results of each step can be checked, and objects are grounded in the image. 
The other method LURE~\cite{zhou2023analyzing} trains a specialized revisor to mask objects with high uncertainty in the descriptions and regenerates the responses again.

\section{Extended Techniques}
\label{sec:tech}

\subsection{Multimodal In-Context Learning}
\label{sec:micl}

ICL is one of the important emergent abilities of LLMs. 
There are two good traits of ICL: 
(1) Different from traditional supervised learning paradigms that learn implicit patterns from abundant data, the crux of ICL is to learn from analogy~\cite{dong2022survey}. Specifically, in the ICL setting, LLMs learn from a few examples along with an optional instruction and extrapolate to new questions, thereby solving complex and unseen tasks in a few-shot manner~\cite{chameleon, mm-react, visprog}. 
(2) ICL is usually implemented in a training-free manner~\cite{dong2022survey} and thus can be flexibly integrated into different frameworks at the inference stage.
A closely related technique to ICL is instruction-tuning (see \S \ref{sec:train_inst_tune}), which is shown empirically to enhance the ICL ability~\cite{wei2021finetuned}.

\begin{table}[!t]
\begin{tcolorbox}

\textcolor[rgb]{0,0.7,0}{<BOS>} Below are some examples and an instruction that describes a task. Write a response that appropriately completes the request \par\medskip

\#\#\# Instruction: \textcolor[rgb]{0,0,0.8}{\{instruction\}}

\#\#\# Image: \textcolor[rgb]{0,0,0.8}{<image>}\ 

\#\#\# Response: \textcolor[rgb]{0.8,0,0}{\{response\}} \par\medskip

\#\#\# Image: \textcolor[rgb]{0,0,0.8}{<image>}\ 

\#\#\# Response: \textcolor[rgb]{0.8,0,0}{\{response\}}

\tcblower 

\#\#\# Image: \textcolor[rgb]{0,0,0.8}{<image>}\ 

\#\#\# Response: \textcolor[rgb]{0.8,0,0}{<EOS>}

\end{tcolorbox}
\caption{A simplified example of the template to structure an M-ICL query, adapted from~\cite{multimodal-gpt}. For illustration, we list two in-context examples and a query divided by a dashed line. 
\textcolor[rgb]{0,0,0.8}{\{instruction\}} and \textcolor[rgb]{0.8,0,0}{\{response\}} are texts from the data sample.  \textcolor[rgb]{0,0,0.8}{<image>} is a placeholder to represent the multimodal input (an image in this case). \textcolor[rgb]{0,0.7,0}{<BOS>} and \textcolor[rgb]{0.8,0,0}{<EOS>} are tokens denoting the start and the end of the input to the LLM, respectively.}
\label{template_ICL}
\end{table}

In the context of MLLM, ICL has been extended to more modalities, leading to Multimodal ICL (M-ICL).
Building upon the setting in (\S \ref{sec:train_inst_tune}), at inference time, M-ICL can be implemented by adding a demonstration set, \ie a set of in-context samples, to the original sample. 
In this case, the template can be extended as illustrated in~\cref{template_ICL}. 
Note that we list two in-context examples for illustration, but the number and the ordering of examples can be flexibly adjusted. In fact, models are commonly sensitive to the arrangement of demonstrations~\cite{dong2022survey, lu2021fantastically}.

\subsubsection{Improvement on ICL capabilities}
Recently, a growing amount of work has focused on enhancing ICL performance under various scenarios. In this section, we trace the development of this field and summarize some relevant works.

MIMIC-IT~\cite{li2023mimic} combines in-context learning with instruction tuning by building an instruction dataset formatted with multimodal context. The model instruction tuned on the introduced dataset shows improved few-shot performance in the caption task.
Emu~\cite{sun2023generative} extends the idea of Flamingo~\cite{alayrac2022flamingo} by introducing extra modalities in model generation and corresponding training corpus. Aided by the introduced vision decoder, \ie Stable Diffusion, the model learns from extra vision supervision and supports more flexibility in output format and in-context reasoning. Specifically, apart from answering in pure text, the model can also give responses in the form of images. 
Sheng~\etal~\cite{sheng2023towards} adopt a similar idea and try to extend output modalities into both text and image. Instead of adopting a specialized encoder for images, the work adopts a unified quantization scheme with a shared embedding layer.

Some other works explore improving few-shot learning performance under specific settings. 
Link-context learning~\cite{tai2023link} focuses on strengthening the causal link between image-label pairs and casts a contrast training scheme by formulating positive and negative image-description pairs.
MMICL~\cite{zhao2023mmicl} aims to augment the capabilities in reasoning with multiple related images. To strengthen the link between image and text, the work proposes a context scheme to transform interleaved image-text data into a uniform format.
Jeong~\cite{jeong2023hijacking} finds that when inserting a small fraction of incoherent images/text as noise, MLLMs can be misled to give responses inconsistent with the context. Based on the observation, the work accordingly proposes a pre-filtering method to remove irrelevant context and facilitate more coherent responses.

\subsubsection{Applications}
In terms of applications in multimodality, M-ICL is mainly used in two scenarios: 
(1) solving various visual reasoning tasks~\cite{few-shot-vqa, alayrac2022flamingo, mm-react, frozen, otter} and (2) teaching LLMs to use external tools~\cite{chameleon, HuggingGPT, visprog}. The former usually involves learning from a few task-specific examples and generalizing to a new but similar question. From the information provided in instructions and demonstrations, LLMs get a sense of what the task is doing and what the output template is and finally generate expected answers. In contrast, examples of tool usage are more fine-grained. They typically comprise a chain of steps that could be sequentially executed to fulfill the task. Thus, the second scenario is closely related to CoT (see \S \ref{sec:mcot}).

\subsection{Multimodal Chain of Thought}
\label{sec:mcot}

As the pioneer work~\cite{wei2022chain} points out, CoT is ``a series of intermediate reasoning steps'', which has been proven to be effective in complex reasoning tasks~\cite{wei2022chain, kojima2022large, zhang2022automatic}.
The main idea of CoT is to prompt LLMs to output not only the final answer but also the reasoning process that leads to the answer, resembling the cognitive process of humans.

Inspired by the success in NLP, multiple works~\cite{vcot, multimodal-cot, vip, cot-prompt-tuning} have been proposed to extend the unimodal CoT to Multimodal CoT (M-CoT). 
We first introduce different paradigms for acquiring the M-CoT ability (\S \ref{sec:mcot_learn}). 
Then, we delineate more specific aspects of M-CoT, including the chain configuration (\S \ref{sec:mcot_config}) and the pattern (\S \ref{sec:mcot_patt}).

\subsubsection{Learning Paradigms}
\label{sec:mcot_learn}
The learning paradigm is also an aspect worth investigating. 
There are broadly three ways to acquire the M-CoT ability, \ie through finetuning and training-free few/zero-shot learning. 
The sample size requirement for the three ways is in descending order. 

Intuitively, the finetuning approach often involves curating specific datasets for M-CoT learning. 
For example, Lu~\etal~\cite{scienceqa} construct a scientific question-answering dataset ScienceQA with lectures and explanations, which can serve as sources of learning CoT reasoning, and finetune the model on this proposed dataset. 
Multimodal-CoT~\cite{multimodal-cot} also uses the ScienceQA benchmark but generates the output in a two-step fashion, \ie the rationale (chain of reasoning steps) and the final answer based on the rationale. 
CoT-PT~\cite{cot-prompt-tuning} learns an implicit chain of reasoning through a combination of prompt tuning and step-specific visual bias.

Compared with finetuning, few/zero-shot learning is more computationally efficient. The main difference between them is that the few-shot learning typically requires hand-crafting some in-context examples so that the model can learn to reason step by step more easily. 
In contrast, the zero-shot learning does not require any specific example for CoT learning. In this case, models learn to use the embedded knowledge and the reasoning ability without explicit guidance by prompting designed instructions like ``Let's think frame by frame'' or ``What happened between these two keyframes''~\cite{vcot, vip}. 
Similarly, some works~\cite{mm-react, visual-chatgpt} prompt models with descriptions of the task and tool usage to decompose complex tasks into sub-tasks.

\subsubsection{Chain Configuration}
\label{sec:mcot_config}
Structure and length are two critical aspects of the reasoning chains.
In terms of structure, current methods can be divided into single-chain and tree-shape methods.
Reasoning with a single chain is a paradigm widely used in various methods~\cite{scienceqa, multimodal-cot}. Specifically, the step-by-step reasoning process forms a single question-rationale-answer chain. 
Recently, some methods have explored using a more complicated scheme, \ie tree-shape chain, for reasoning. Specifically, DDCoT~\cite{zheng2023ddcot} breaks down a question into multiple sub-questions, each of which is solved by LLM itself or visual experts to generate rationales. Then the LLM aggregates and reasons with the rationales to form the final answer.
With respect for chain length, it can be categorized into adaptive and pre-defined formations. 
The former configuration requires LLMs to decide on their own when to halt the reasoning chains~\cite{chameleon, scienceqa, mm-react, multimodal-cot, visual-chatgpt, visprog}, while the latter setting stops the chains with a pre-defined length~\cite{cot-prompt-tuning, CAT, vip, vcot}.

\vspace{-0.5em}
\subsubsection{Generation Patterns}
\label{sec:mcot_patt}
How the chain is constructed is a question worth studying. 
We summarize the current works into (1) an infilling-based pattern and (2) a predicting-based pattern. Specifically, the infilling-based pattern demands deducing steps between surrounding context (previous and following steps) to fill the logical gaps~\cite{vip, vcot}. 
In contrast, the predicting-based pattern requires extending the reasoning chains given conditions such as instructions and previous reasoning history~\cite{chameleon, scienceqa, mm-react, multimodal-cot, visual-chatgpt, visprog}. 
The two types of patterns share a requirement that the generated steps should be consistent and correct.

\subsection{LLM-Aided Visual Reasoning}
\label{sec:vr}

\subsubsection{Introduction}
\label{sec:vr_intro}

Inspired by the success of tool-augmented LLMs~\cite{parisi2022talm, gao2022pal, schick2023toolformer, nakano2021webgpt}, some researches have explored the possibilities of invoking external tools~\cite{gpt4tools, visprog, mm-react, chameleon} or vision foundation models~\cite{mm-react, CAT, chatcaptioner, sm, visual-chatgpt, idealgpt,udandarao2022sus} for visual reasoning tasks. 
Taking LLMs as helpers with different roles, these works build task-specific~\cite{PointCLIP-V2, CaFo, CAT} or general-purpose~\cite{HuggingGPT, mm-react, visual-chatgpt, chameleon, visprog} visual reasoning systems.

Compared with conventional visual reasoning models~\cite{anderson2018bottom, yu2019deep, gao2019dynamic}, these works manifest several good traits: (1) Strong generalization abilities. Equipped with rich open-world knowledge learned from large-scale pretraining, these systems can easily generalize to unseen objects or concepts with remarkable zero/few-shot performance~\cite{chameleon, visprog, idealgpt, PointCLIP-V2, CaFo, SMs}. 
(2) Emergent abilities. Aided by strong reasoning abilities of LLMs, these systems can perform complex tasks. For example, given an image, MM-REACT~\cite{mm-react} can interpret the meaning beneath the surface, such as explaining why a meme is funny.
(3) Better interactivity and control. Traditional models typically allow a limited set of control mechanisms and often entail expensive curated datasets~\cite{gan2017stylenet, mathews2016senticap}. 
In contrast, LLM-based systems have the ability to make fine control in a user-friendly interface (\eg click and natural language queries)~\cite{CAT}.

For this part, we start with introducing different training paradigms employed in the construction of LLM-Aided Visual Reasoning systems (\S \ref{sec:vr_train}). 
Then, we delve into the primary roles that LLMs play within these systems (\S \ref{sec:vr_func}).

\subsubsection{Training Paradigms}
\label{sec:vr_train}
According to training paradigms, LLM-Aided Visual Reasoning systems can be divided into two types, \ie training-free and finetuning.

\noindent \textbf{Training-free.}
With abundant prior knowledge stored in pre-trained LLMs, an intuitive and simple way is to freeze pre-trained models and directly prompt LLMs to fulfill various needs. 
According to the setting, the reasoning systems can be further categorized into few-shot models~\cite{chameleon, HuggingGPT, mm-react, visprog} and zero-shot models~\cite{PointCLIP-V2,CAT}. 
The few-shot models entail a few hand-crafted in-context samples (see \S \ref{sec:micl}) to guide LLMs to generate a program or a sequence of execution steps. 
These programs or execution steps serve as instructions for corresponding foundation models or external tools/modules. 
The zero-shot models take a step further by directly utilizing LLMs' linguistics/semantics knowledge or reasoning abilities. 
For example, PointCLIP V2~\cite{PointCLIP-V2} prompts GPT-3 to generate descriptions with 3D-related semantics for better alignment with corresponding images. 
In CAT~\cite{CAT}, LLMs are instructed to refine the captions according to user queries.

\noindent \textbf{Finetuning.}
Some works adopt further finetuning to improve the planning abilities with respect to tool usage~\cite{gpt4tools} or to improve localization capabilities~\cite{lai2023lisa,wu2023textit} of the system. For example, GPT4Tools~\cite{gpt4tools} introduces the instruction-tuning approach (see \S \ref{sec:train_inst_tune}). 
Accordingly, a new tool-related instruction dataset is collected and used to finetune the model.

\subsubsection{Functions}
\label{sec:vr_func}
In order to further inspect what roles LLMs exactly play in LLM-Aided Visual Reasoning systems, existing related works are divided into three types:
\begin{itemize}
    \item LLM as a Controller
    \item LLM as a Decision Maker
    \item LLM as a Semantics Refiner
\end{itemize}

The first two roles are related to CoT (see \S \ref{sec:mcot}). It is frequently used because complex tasks need to be broken down into intermediate simpler steps. 
When LLMs act as controllers, the systems often finish the task in a single round, while multi-round is more common in the case of the decision maker.
We delineate how LLMs serve these roles in the following parts.

\noindent \textbf{LLM as a Controller.}
In this case, LLMs act as a central controller that (1) breaks down a complex task into simpler sub-tasks/steps and (2) assigns these tasks to appropriate tools/modules. The first step is often finished by leveraging the CoT ability of LLMs. Specifically, LLMs are prompted explicitly to output task planning~\cite{HuggingGPT} or, more directly, the modules to call~\cite{chameleon, gpt4tools, visprog}. For example, VisProg~\cite{visprog} prompts GPT-3 to output a visual program, where each program line invokes a module to perform a sub-task.
In addition, LLMs are required to output argument names for the module input. 
To handle these complex requirements, some hand-crafted in-context examples are used as references~\cite{chameleon, HuggingGPT, visprog}.
This is closely related to the optimization of reasoning chains (see \S \ref{sec:mcot}), or more specifically, the least-to-most prompting~\cite{zhou2022least} technique.
In this way, complex problems are broken down into sub-problems that are solved sequentially.

\noindent \textbf{LLM as a Decision Maker.}
In this case, complex tasks are solved in a multi-round manner, often in an iterative way~\cite{idealgpt}. 
Decision-makers often fulfill the following responsibilities: 
(1) Summarize the current context and the history information, and decide if the information available at the current step is sufficient to answer the question or complete the task; (2) Organize and summarize the answer to present it in a user-friendly way.

\noindent \textbf{LLM as a Semantics Refiner.}
When LLM is used as a Semantics Refiner, researchers mainly utilize its rich linguistics and semantics knowledge. 
Specifically, LLMs are often instructed to integrate information into consistent and fluent natural language sentences~\cite{SMs} or generate texts according to different specific needs~\cite{CAT, PointCLIP-V2, CaFo}.

\section{Challenges and Future Directions}
\label{sec:future}

The development of MLLMs is still in a rudimentary stage and thus leaves much room for improvement, which we summarize below:

\begin{itemize}

\item Current MLLMs are limited in processing multimodal information of long context. This restricts the development of advanced models with more multimodal tokens, \eg long-video understanding, and long documents interleaved with images and text.

\item MLLMs should be upgraded to follow more complicated instructions. For example, a mainstream approach to generating high-quality question-answer pair data is still prompting closed-source GPT-4V because of its advanced instruction-following capabilities, while other models generally fail to achieve.

\item There is still a large space for improvement in techniques like M-ICL and M-CoT. Current research on the two techniques is still rudimentary, and the related capabilities of MLLMs are weak. Thus, explorations of the underlying mechanisms and potential improvement are promising.

\item Developing embodied agents based on MLLMs is a heated topic. It would be meaningful to develop such agents that can interact with the real world. Such endeavors require models with critical capabilities, including perception, reasoning, planning, and execution. 

\item Safety issues. Similar to LLMs, MLLMs can be vulnerable to crafted attacks~\cite{jeong2023hijacking,zhao2023evaluating,shayegani2023jailbreak}. In other words, MLLMs can be misled to output biased or undesirable responses. Thus, improving model safety will be an important topic.

\end{itemize}

\section{Conclusion}
\label{sec:conclusion}

In this paper, we perform a survey of the existing MLLM literature and offer a broad view of its main directions, including the basic recipe and related extensions. Moreover, we underscore the current research gaps that need to be filled and point out some promising research directions. We hope this survey can offer readers a clear picture of the current progress of MLLM and inspire more works.

\footnotesize
\bibliographystyle{IEEEtran}
\bibliography{IEEEabrv, references}

\end{document}